# An Artificial Intelligence Approach for Interpreting Creative Combinational Designs

Liuqing Chen [a], Shuhong Xiao [a], Yunnong Chen [a], Linyun Sun [a], Peter R.N. Childs [b], Ji Han [c, *]

[a] *Department of Computer Science and Technology, Zhejiang University, Hangzhou 310030, China;*

[b] *Dyson School of Design Engineering, Imperial College London, London SW7 2AZ, UK;*

[c] *University of Exeter Business School, London, SE1 7TY, UK*

*Corresponding author: j.han2@exeter.ac.uk*

# An Artificial Intelligence Approach for Interpreting Creative Combinational Designs

Combinational creativity, a form of creativity involving the blending of familiar ideas, is pivotal in design innovation. While most research focuses on how combinational creativity in design is achieved through blending elements, this study focuses on the computational interpretation, specifically identifying the 'base' and 'additive' components that constitute a creative design. To achieve this goal, the authors propose a heuristic algorithm integrating computer vision and natural language processing technologies, and implement multiple approaches based on both discriminative and generative artificial intelligence architectures. A comprehensive evaluation was conducted on a dataset created for studying combinational creativity. Among the implementations of the proposed algorithm, the most effective approach demonstrated a high accuracy in interpretation, achieving 87.5% for identifying 'base' and 80% for 'additive'. We conduct a modular analysis and an ablation experiment to assess the performance of each part in our implementations. Additionally, the study includes an analysis of error cases and bottleneck issues, providing critical insights into the limitations and challenges inherent in the computational interpretation of creative designs.

# 1 Introduction

Combinational creativity is the easiest form of creativity for human beings among the three types of creativity (exploratory, transformational, and combinational) proposed by Boden (1996). It involves blending novel combinations of familiar ideas, which is achieved by connecting ideas that were previously unrelated. A number of people have explained creativity by using the term 'combinational creativity'. For example, Frigotto and Riccaboni (2011) described that the nature of creativity is to combine; Henriksen et al. (2014) suggested that creativity is the process of creating something new by generating new combinations and alterations with existing ideas; Childs (2018) indicated that combinations of essential mental capabilities lead to creativity; and Sawyer et al. (2024) emphasize that combination is one of the most important ways to explain creativity. Combinational creativity has been employed widely in design through various forms, such as bisociation which connects unrelated and often conflicting ideas in new ways (Koestler, 1964), and analogy exploring shared conceptual space (Boden, 2009).

In practice, combinational ideas can be developed by associating diverse elements, including words, ideas, concepts, images, and even musical styles and artistic genres (Ward and Kolomyts, 2010). In the context of this study, we specifically focus on the conventional form of combinational creativity known as noun-noun combination. Here, noun refers to both single noun words like 'pencil' or phrases like 'mechanical pencil'. The first noun is also known as the 'base', signifying it as the foundational element in the formation of the creativity design, while the second is termed the 'additive', representing the supplementary part that enhances the design (Han et al, 2018). Researchers in the field, such as Nagai et al. (2009) and Ward et al. (2013), have delved into the realm of noun-noun combinations, examining their intricacies and associated interpretations. For instance, Nagai et al. (2009)

illustrated the use of three compound phrase interpretation methods for generating fresh concepts, including property mapping, concept blending, and integration.

In recent years, there is an increasing interest in employing combinational creativity. Most literature focuses on the integrative process of 'combination', aimed at aiding designers in generating new ideas during the early stages of the design process. For instance, Bacciotti et al. (2016) introduced a computational method that combines concepts from different dimensions to identify scenarios that stimulate creative idea generation. Georgiev et al. (2017) synthesized scenes from various contexts, thereby encouraging the creation of new design ideas. In addition to textual representations of ideas, Han et al. (2018) developed 'the Combinator', which provided a visual expression of creativity through the blending of the original concept images. Utilizing generative adversarial networks, Chen et al. (2019) achieved more harmonious outputs of creative images through artificial intelligence.

As an emerging branch of computational creativity, data-driven creative methods are being increasingly utilized (Kelly et al, 2015). In the field of combinational creativity, although a wealth of creative cases in graphic and textual forms can be found on the internet, structuring these resources for data-driven design proves challenging, as it requires expertise and specific physical environments (Han et al, 2019). Moreover, the attempt to automate the deconstruction process of combinational creativity, particularly in terms of interpreting 'base' and 'additive' elements, is still absent. It could be elusive for machines to understand the rationale and mechanism behind such combinations (Boden, 2009). Addressing this aspect is crucial as it offers valuable insights into the data-driven design cycles for design creativity (Chen, 2020). It benefits knowledge management in conceptual design by extracting structured design concepts from existing design information, facilitating the reuse of design knowledge to accelerate future designs. Besides, it enables the assessment of creative products from an original concept perspective.

In this study, our motivation is to fill the gap in the interpretative process of combinational creativity. We aim to guide this process through a semi-automated approach that does not require extensive knowledge in the field of design. Inspired by the three driving forces identified by Han et al. (2019), we propose a computational algorithm that employs advanced artificial intelligence techniques for interpreting creative combinational designs. Artificial intelligence technologies have been widely utilized in design practices. Generally, they do not address complex issues from a holistic system perspective as human designers do, but instead manage complexity through the continuous iteration of simple tasks (Roberto et al, 2020). In our context, 'interpreting' refers to the process of acquiring meta-knowledge about design. This involves identifying the combination pairs — namely, the 'base' and 'additive' elements — that constitute the essence of combinational creativity. Specifically, we proposed a heuristic algorithm that breaks down the interpretative process into multiple simple tasks, integrating both computer vision and natural language processing technologies. We implemented this algorithm based on multiple approaches, including both discriminative and generative artificial intelligence. Our approaches were rigorously tested on a dataset (Han et al, 2019) specifically curated for combinational creativity. Impressively, the most effective method demonstrated a high recognition accuracy, achieving 87.5% for identifying 'base' elements and 80% for 'additive' elements. Furthermore, we established baselines using generative large language models (LLMs) for comparison. The results indicate that our algorithm significantly enhances the identification of combination pairs. We present the following contributions:

(1) This is the first study that proposed a computational method for interpreting combinational creativity. It fills a crucial gap in data-driven design cycles by transforming design creativity into meta-data, thus enhancing the understanding of creativity processes in design.

(2) We developed a heuristic-based interpretation algorithm, grounded in an understanding of how designs are formed through combinational creativity. This algorithm, integrating computer vision and natural language processing technologies, was implemented across various discriminative and generative AI architectures.

(3) Our approaches were validated on a dataset of combinational creativity, showing promising predictive performance. We conducted a modular analysis of the discriminative AI-based approaches, with a discussion about the performance and potential issues of each component. Furthermore, by contrasting with baselines, we demonstrated the effectiveness of our approach, underlining its viability and robustness in interpreting creative designs.

## 2 Related works

### *2. 1 Artificial Intelligence in Combinational Creativity*

In recent years, data-driven artificial intelligence technologies have been recognized as capable of engaging in creative tasks like humans (Wang et al, 2024; Zhou et al, 2024). In this section, we review the applications of AI in the field of combinational creativity and explain how they inspire this study. Generally, most previous work has focused on the generative and subsequent evaluative phases of creativity, using existing structured creative data as input to produce creative works and perspectives. This study, however, concentrates on the interpretation side, aiming to achieve a reverse transformation back to data. Typically, a complete data-driven learning process includes these two opposing branches (Chen, 2020).

In this study, based on the methodology of using artificial intelligence to handle complex design tasks through the iteration of simple steps (Roberto et al, 2020), a straightforward step in our algorithm involves extracting all nouns from textual descriptions as potential candidates, particularly considering the characteristics of 'base' and 'additive' as nouns. However, the combinations of 'base' and 'additive' are typically complex and varied, involving harmonious interactions between concepts and the specialized insights of designers. Previous research by Han et al. (2017; 2019) provided a critical foundation for understanding the relationships. Specifically, in collaboration with experienced designers, they identified three representative ways of combining 'base' and 'additive', termed as the problem-driven, similarity-driven, and inspiration-driven approaches, shown as Figure 1. Other works, such as 'the Combinator' and the design GAN proposed by Chen et al., incorporate both textual and visual data. Through cross-modal computational methods, they offer new perspectives for understanding the creative process. Other works, such as 'the Combinator' (Han et al, 2018) and the design GAN proposed by Chen et al. (2019), incorporate both textual and visual data. Through cross-modal computational methods, they offer new perspectives for understanding the creative process. Recently, the emergence of generative artificial intelligence (GAI) has also sparked widespread discussion in the fields of design and creativity. With the extensive integration of design corpora from the natural world, GAI has shown unique advantages in understanding complex design concepts (Franceschelli et al, 2023). Furthermore, prompting allows designers to input additional insights and ways of thinking, aiding the creative process in a manner more aligned with human habits (Di Fede et al, 2022). This has been widely utilized in areas such as conceptual design (Ma et al, 2023; Wang et al, 2023) and product interaction design (Friedl et al, 2023).

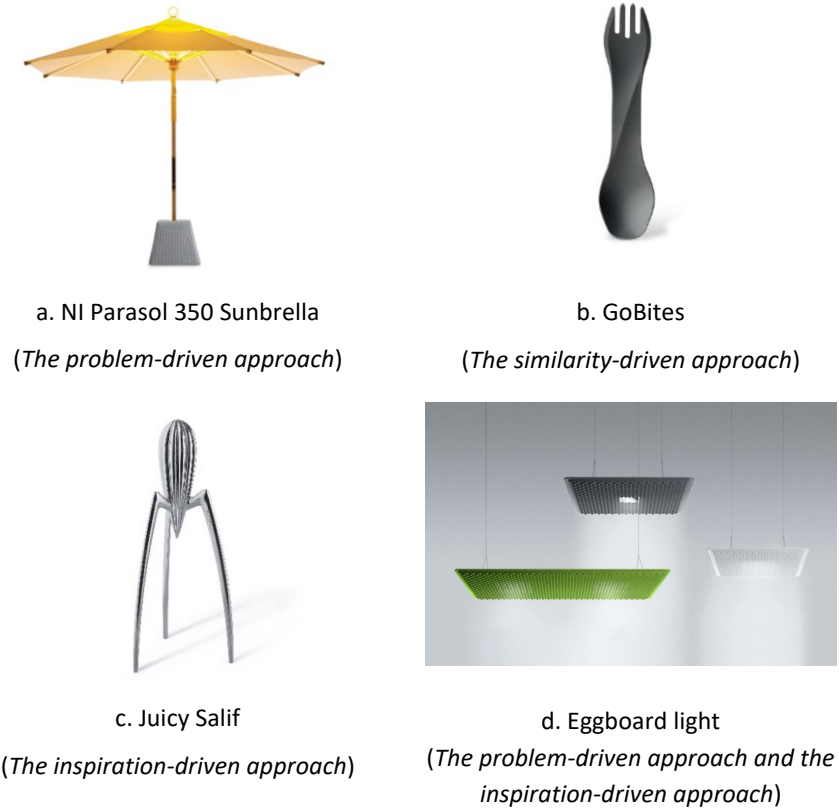

a. NI Parasol 350 Sunbrella  
(*The problem-driven approach*)

b. GoBites  
(*The similarity-driven approach*)

c. Juicy Salif  
(*The inspiration-driven approach*)

d. Eggboard light  
(*The problem-driven approach and the inspiration-driven approach*)

Figure 1. Examples of the three combinational creativity driven approach (Han et al., 2017)

## *2. 2 From design creativity to data*

Benefiting from the active community of computational creativity (Colton et al, 2015), the research on creative systems has achieved much in different aspects, such as framework (Carnovalini et al, 2020), computational creativity models (Colton and Wiggins, 2012; Marrone et al, 2022), and related applications (Cook and Colton, 2018; Colton et al, 2021; Oppenlaender, 2022). Unlike most studies focused on the integrative processes of creativity, this study concentrates on the interpretation side, aiming to achieve a reverse transformation from creativity to data (Chen, 2020). Technically, this work relies on data mining to delve into creativity. In this section, we provide a brief review of data mining in the field of creativity, outlining the key techniques and their applications in enhancing our understanding of creative processes.

Design creativity exists in a variety of formats, of which textual and image are the two widely used digital formats. In design engineering, textual data are analysed for various purposes. For example, Chaklader and Parkinson (2017) analyse consumer reviews to provide information quickly and economically for the establishment of design specifications related to human-artifact interaction. Song and Luo (2019) applied patent mining techniques to search for precedents of a product design in patent databases in order to learn about relevant prior arts, seek design inspiration, or for benchmark purposes. The capabilities of natural language processing (NLP) methods in handling unstructured text make them a crucial tool in design research (Siddharth et al, 2022), such as knowledge reuse (Li et al, 2021), needs elicitation (Lin et al, 2012) and biomimicry (Arslan et al, 2022).

Image data, including sketches, drawings, product sample images, and CAD designs, are also widely used in design engineering. From the perspective of engineering, image data mainly expresses product's functionality and behaviour, and manufacturing procedure. On the other hand, they tend to illustrate product shape, appearance, and visual feelings from the design perspective. In this case, a significant amount of research in data-driven design leverages image data to facilitate and enhance the design process, as well as to foster innovation in design. For example, Dering and Tucker (2017) introduced a deep learning method using 3D convolutions. This approach efficiently predicts functional aspects, like seating, liquid storage, and sound emission, in digital designs. While Wang et al. (2017) developed an image mining algorithm that yields insights into shape variability and enables the creation of accurate 3D models.

## 3 Approach overview

### *3.1 Problem statement*

In this study, we consider a noun-noun combinational creative designs as comprising a primary base idea and an additive idea. While complex designs may have multiple bases or additives, our focus is on single pairs to simplify the modelling of combinational creativity. Notably, no existing literature, to the authors' knowledge, delves into the computational interpretation of these designs. It thereby raises a research challenge that how the base and additive could be computationally extracted respectively, when a combinational design is provided.

Combinational designs could be expressed or presented in various digital formats, involving images, texts, and even three-dimensional models. Image and text formats are the ones used most often, as they are commonly used in nowadays digital systems. Therefore, our objective is to automatically determine the base and additive components of combinational design products, from provided images and textual descriptions, with a computational approach. Using Figure 1(d) as an instance, from the image of the product and its textual description below:

*"The design of the Eggboard **pendant luminaire** picks up this principle, translating it into a high-quality lighting option. Surfaces of simple **egg cartons** possess outstanding sound absorption qualities thanks to the specific surface structure."* (Eggboard, 2016)

It is expected that 'pendant luminaire' can be extracted as the base, and the phrase 'egg cartons' can be extracted as the additive of the combinational design 'Eggboard', along with the image interpretation from Figure1(d). In this example, our initial challenge is to identify key elements like 'pendant luminaire' and 'egg cartons' among multitude of noun entities present in the design description and imagery. The linguistic expression of creativity and metaphors (Han et al, 2019), which is widely employed in these descriptions, complicates the task of accurately extracting the specific nouns we need from the text. Regarding design image interpretation, the 'base' and 'additive' elements of a product are typically merged into a singular physical form. This integration is often accompanied by transformations and distortions of explicit traits such as shape, texture, size, and materials, further complicating the task of visually distinguishing these components within the design. The second challenge involves distinguishing the roles of the two extracted nouns as either 'base' or 'additive'. This demands not only an efficient text parsing capability from an AI model but also a profound understanding of the intrinsic connections between design concepts.

*3.2 The Algorithm for interpreting combinational creative designs*

---

**Algorithm 1: The Algorithm for interpreting combinational creative designs**

---

   **Input:** the product image $I$, the product description $T$
   **Output:** Base element, Additive element
**BEGIN**
  // Step 1: Identify Base Element
  Base = ImageInterpretation ($I$)
  // Step 2: Extract all Nouns or Noun Phrases as Potential Additives
  PotentialAdditives = NounEntityExtraction ($T$)
  // Step 3: Access Potential Additives with Relation
  **FOR** each AdditiveCandidate in PotentialAdditives **DO**
      Relation = CheckRelation ($T$, Base, AdditiveCandidate)
         **IF** RelationMatch (Relation, PredefinedRelation) **Then**
            Additive = AdditiveCandidate
        **END IF**
  **END FOR**
  **RETURN** Base, Additive
**END**

---

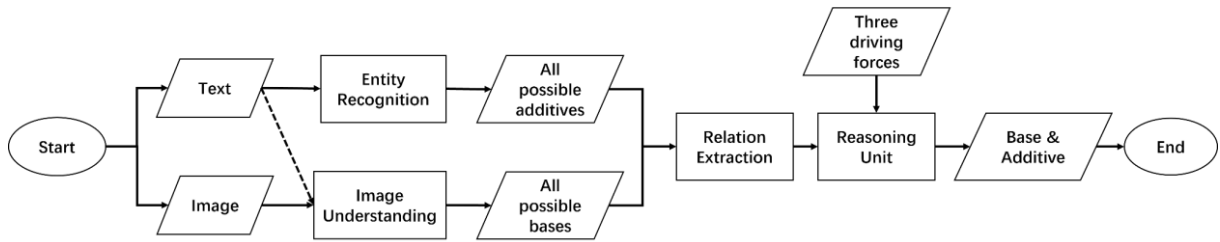

Figure 2. Flowchart of the proposed framework

In order to interpret a combinational creative design, alternatively, to extract the 'base' and 'additive' pair from the corresponding image and textual description of the design, an integrated interpretation algorithm is proposed in this study. We present the overall flowchart in Figure 2 and a pseudo code shown as Algorithm 1. Given the product image $I$ and textual description $T$, we start with the image interpretation. By understanding the main subject in the image, we can narrow down the numerous nouns to focus on the potential 'base' and 'additive' elements, as they can be explicitly manifested through factors such as shape and appearance. It is intuitive to observe from the examples in Figure 1, the 'base' (foundational part) is usually more discernible. This is attributed to the fact that the base constitutes the product's principal structure or core functionality, thereby delineating its essential features and intended uses. Regarding the 'additive' aspect, based on the three combination strategies mentioned, it often serves as a functional expansion or a creative supplement. Typically, it is more challenging to discern due to its nuanced and integrative nature within the product. In this case, we use the image interpretation results as a reliable prediction of the 'base' and serve as additional information that aids in determining the 'additive' elements.

To identify the 'additive' elements, we first leverage the characteristic of noun-noun combinations, extracting all nouns or noun phrases from product descriptions (textual) as potential candidates for the additive component as step 2 in Algorithm 1. For the 'base' and each potential additive, we assess whether the latter is indeed an additive by understanding their relationship in the context of the description. Based on the characteristics of combinational creativity, the 'base' and 'additive' components are always combined together based on specific principles, such as the three classical forms discussed in Section 2.1. Therefore, we can identify them through their specific relationships. Through scrutinizing the three common types of combinatorial creativity, we summarized several predefined relations, as shown in Table 1, to assist in determining additives. For the problem-driven approach, common relational terms include 'solution', indicating that the additive provides a specific solution to the base; and 'integration', suggesting a combination of the additive with the base to solve a more complex problem. In the similarity-driven approach, the term 'complementarity' reflects how the additive complements the base, enhancing its original characteristics or functionalities, while 'harmonization' denotes a harmonious combination of the two in function or design, improving overall consistency and effectiveness. For the inspiration-driven approach, 'innovation' indicates that the additive brings novel and unique features or functionalities to the base, and 'transformation' implies that the additive completely changes the traditional use or appearance of the base. For each relation term, their semantically similar substitutes are equally valid. For instance, 'integration' and 'part of' are interchangeable in context. These summarized relations serve as PredefinedRelation in the Algorithm to help identify the relation between base and additive.

Table 1. An overview of relationship between base and additive

| Combinational creativity approach | Relation | Description |
|---|---|---|
| Problem-Driven | Solution | The additive provides a specific solution to the base. |
| | Integration | The additive combines with the base to solve a more complex problem. |
| Similarity-Driven | Complementarity | The additive complements the base, enhancing its original characteristics or functionalities. |
| | Harmonization | The additive and base harmoniously combine in function or design, improving overall consistency and effectiveness. |
| Inspiration-Driven | Innovation | The additive brings novel and unique features or functionalities to the base. |
| | Transformation | The additive completely changes the traditional use or appearance of the base. |

**4 Implementation**

In this section, we begin by introducing the dataset used in our study. To operationalize our algorithm, we employ a trio of modules: an image recognition module for image interpretation, an entity recognition module for extracting nouns, and a relation extraction module for checking relationships. Since each module exclusively processes either image or text data, this approach is termed a unimodal method. Additionally, we have made efforts to integrate both images and textual data in each module, thus achieving a multimodal approach. We refer to both implementations as discriminative methods, primarily because they rely on discriminative AI models designed for classification tasks. On the other hand, we have also developed an approach based on generative AI, attempting to guide the inference of large language models (LLMs) through our algorithm. Given the inherent proficiency of LLMs in image-text reasoning tasks, we aim to ascertain whether our algorithm can enhance their intrinsic ability to interpret combinational creativity. For this purpose, we have also implemented a vanilla version as a baseline for comparison, allowing us to evaluate the effectiveness of our algorithmic intervention.

*4.1 Dataset*

This study employs the dataset[1] developed by Han et al. (2019), which is specifically curated for investigating the driven approaches of combinational creativity in design. The dataset encompasses data on two hundred products originating from combinational creativity, including their names, images, and descriptions. These products were meticulously selected from the award winners of prestigious design competitions, such as the iF and Red Dot design awards, with their detailed information sourced directly from the competitions' official websites. As detailed in Table 2, for each product sample, the dataset limits the representation to one image and a maximum of five sentences in the textual description. A team of design experts analysed the 200 samples, identifying the 'base' and 'additive' elements of each product, which were then incorporated into the dataset. This extraction process was performed manually and subsequently validated by the experts. For instance, in the case of sample 2 'Sharp 1', the 'base' is identified as a 'knife block', and the 'additive' as a 'knife sharpener', deduced from its textual and visual description. It is important to note that the terms used to describe the 'base' and 'additive' in the dataset directly correspond to the language found in the product descriptions and/or names. For example, the 'base' for sample 1, a 'drying rack', was not explicitly mentioned in the product's description but was inferred from its name. This careful approach ensures the dataset's integrity in accurately representing the elements of combinational design.

---
[1] https://zenodo.org/records/11044248

Table 2. An overview of computational design creativity dataset

| No. | Name | Image | Description | Base | Additive |
|---|---|---|---|---|---|
| 1 | Baby Bottle Drying Rack | 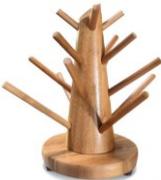 | The form is inspired by a natural tree shape and eliminates water pooling and prevents minerals and bacteria from building up. | Drying Rack | Tree |
| 2 | Sharp 1 | 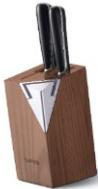 | This knife block set and its integrated knife sharpener are a space-saving combination of different functions. It saves users from having to search for a knife sharpener when needed. | Knife Block | Knife Sharpener |

| 3 | Origami | 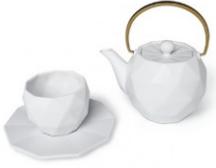 | Inspired by the origami paper folding technique, the surface of this teaware features an inventive structure. It lends the two-piece tea set a unique feel while also adopting the round shape of classic tea ware. | Tea set | Origami Paper Folding |
| ... | ... | ... | ... | ... | ... |

## 4.2 The discriminative approaches

### 4.2.1 A unimodal method

In the field of deep learning, computer vision (CV) and natural language processing (NLP) are the two distinct disciplines characterized by different types of input modalities. In this section, we utilize advanced computer vision and natural language processing technologies to achieve interpreting of combinational pairs. Here, unimodal refers to the concept that each component we propose processes only a single type of input, namely images or text. Given an input image, image classification identifies the subject present in the image. Generally, the types of objects that an image model can 'recognize' are limited and depend on the characteristics of the dataset used to train the model. For instance, ImageNet (Deng et al, 2019) contains over 14 million images labeled with over 20,000 categories, ranging from everyday objects to animals and landscapes. Another dataset, COCO (Lin et al, 2014), features over 330,000 images with 80 object categories, including people, vehicles, and street scenes. Typically, combinational design products are centered around everyday items, like the drying racks, knife sharpeners, and teaware shown in Table 2. These product categories can generally be found within the 1000 classes of ImageNet. Considering the potential limitations in class variety within the ImageNet-1000 dataset, particularly concerning the categories relevant to combinational design creativity, this study opts for a commercial image prediction API by Clarifai[2], which encompasses a broader range of categories, as a substitute for models trained on ImageNet. This is to ensure that all bases of the combinational designs are predictable.

In order to extract entities and relations from textual descriptions, which also include the names of the products, two different models were proposed to perform the two extractions, respectively. To extract all possible noun entities at sentence-level, we utilized the named entity recognition (NER) module of spaCy (Honnibal et al, 2017) in this study due to its excellent performance and popularity. To determine which noun aligns with the base result from image recognition, we employed spaCy's similarity check, assessing semantic similarity. Our approach to discerning relationships between two entities involves the use of Relation Extraction (RE) techniques. Typically, RE processes involve inputting textual descriptions and two target entities, from which the system deduces a contextually based relational interpretation. Often, texts contain additional entities and their interrelations, known as contextual relations. Although these are not the primary focus, they can significantly influence the interpretation of the target relation. Thereby, this study has adopted a context-aware architecture to extract the target relation (Sorokin and Gurevych, 2017).

---

[2] Clarifai official website: https://www.clarifai.com/

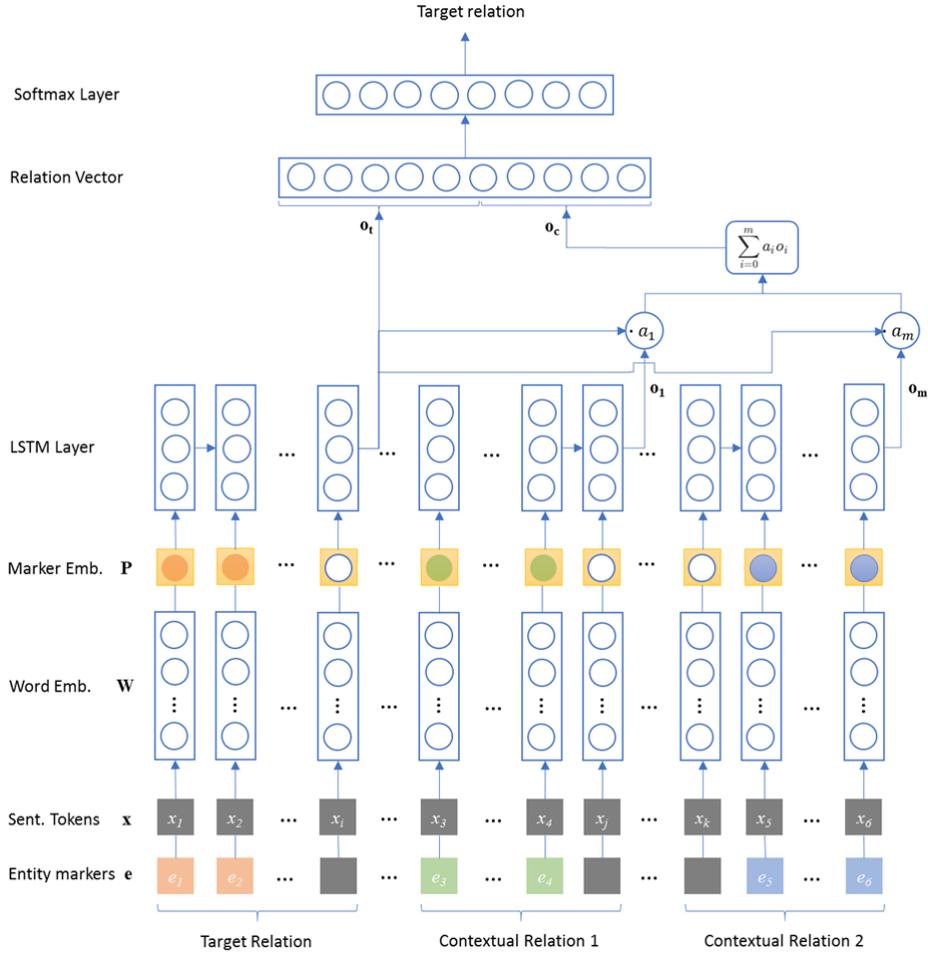

Figure 3. The architecture of the context-aware RE model

As depicted in Figure 3, the RE model initiates by segmenting the description into a series of tokens, x = {x1, x2, …, xn}, using whitespace as the delimiter. When considering nouns as possible additive candidates, we individually examined their relationships with the base. Here, the interaction between the base and a specific noun is considered the target relation, while those between the base and other nouns are treated as contextual relations. Entities markers are implemented to classify each token as belonging to an entity or not. Subsequently, the model maps the token sentence to a k-dimensional embedding vector employing a matrix $W \in \mathbb{R}^{|V| \times k}$. In the matrix, |V| refers to the size of the vocabulary. Here, a pre-trained Word2Vec model from Google by Mikolov et al. (2013), in which three million 300-dimensional embeddings of words or phrases are trained, is employed. Similarly, by randomly initializing each entity, the entity markers are converted into a marker embedding matrix $P \in \mathbb{R}^{3 \times d}$. In the matrix, d refers to the dimension of the embedding, of which there are three marker types. Each marker embedding is concatenated with word embedding $(W_n, P_n)$ and fed into the LSTM layer. The output is depicted as $O_i \in \mathbb{R}^O$ for contextual relation and $O_t \in \mathbb{R}^O$ for target relation. An attention mechanism is implemented in the mode, of which a score is computed for a contextual relation regarding to the target relation: $g(o_i, o_s) = o_i A o_s$ (A refers to a weight matrix learned in the LSTM layer). A weight can then be calculated by the following equation (1):

$$a_i = \frac{\exp(g(o_i, o_t))}{\sum_{j=0}^{m} \exp(g(o_j, o_t))} \quad (1)$$

The contextual relation representations are summed up by the following equation (2):

$$o_c = \sum_{i=0}^{m} a_i o_i \quad (2)$$

The context representation $O_c$ is concatenated with the target relation: $O = [O_t, O_c]$. The concatenated vector is then fed into the softmax layer for predicting the type of target relation. The context-aware RE model was trained by using the Wikidata dataset (Sorokin and Gurevych, 2017), which involves 284,295 relation triples and 578,99 relation instances for training, as well as supports 353 different relation types. Finally, we utilized spaCy's similarity check to assess the similarity between the relation identified and the key terms predefined in Table 2. The noun source with the highest similarity was then reported as our additive.

*4.2.2 A multimodal method*

Building on the unimodal implementation, a natural progression is to consider extending each module to a multimodal approach, to see if it yields better results. In this section, we demonstrate how to incorporate both images and textual descriptions into the identification process of the base and the additive. To identify the base from the textual description, our approach involves transforming both the text and the image into a joint representational space. We then used similarity as the criterion to find the target base element. There are various foundations that can assist us in achieving this objective. For instance, CLIP (Radford et al, 2021), ALIGN (Jia et al, 2021), and Imagebert (Qi et al, 2020) all effectively perform joint representation of text and images. In this study, we selected Contrastive Language-Image Pre-Training (CLIP) as our joint representation model, as it is readily accessible and continuously undergoes updates[3]. CLIP is a neural network trained on more than 1.28 million (image, text) pairs and is commonly used for aligning and transforming text and images (Saharia et al, 2022; Zhang et al, 2023). We first utilized the NER (named entity recognition) module of spaCy to find all noun entities as described in 4.2.1. Given each noun $N_i$ as base candidate and product image $I$, CLIP converts both into high-dimensional vectors, $V(N_i)$ and $V(I) \in \mathbb{R}^D$, respectively, through its dual encoding mechanism. We then calculated the compatibility score $S(N_i, I)$ between each noun $N_i$ and image $I$ with cosine similarity by the following equation (3):

$$S(N_i, I) = \frac{V(N_i) \cdot V(I)}{\|V(N_i)\| \, \|V(I)\|} \qquad (3)$$

The noun $N_{base}$ that yields the highest compatibility score with the image $I$ is considered the base element. Mathematically, it can be expressed as:

$$N_{base} = argmax_{N_i}(S(N_i, I)) \qquad (4)$$

---

[3] https://openai.com/research/clip

For interpreting the additive, we employed a multimodal relation extraction model, which enhances the understanding of potential semantic gaps in the sentence by incorporating the visual modality. In this study, we used MEGA (Zheng et al, 2021) as our relation extraction model. It utilizes object detection technology to extract potential objects from images and form a scene graph, serving as a complement to the textual semantics. We adapted the MEGA framework for our specific application. Recognizing that in scenarios involving the prediction of creative combinational pairs, the base and the additive often correspond to different aspects of the same object within an image. In this case, we opted to exclude the original multimodal graph structural alignment module from MEGA. This decision was informed by the understanding that such structured alignment may not be conducive to extracting meaningful information in cases where the base and additive are intrinsically linked within a single object's representation.

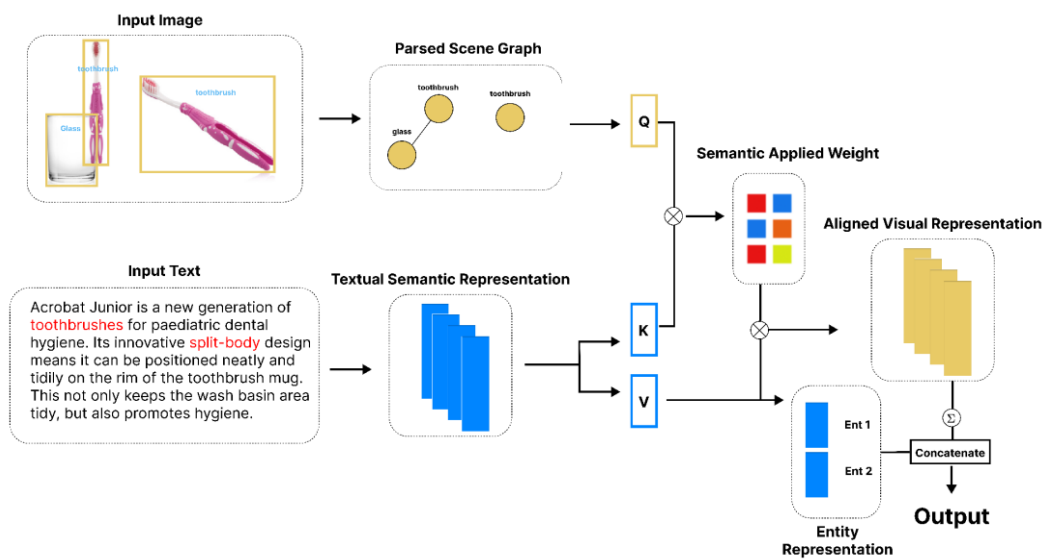

Figure 4. The architecture of the MEGA model

Figure 4 illustrates the framework of the model after adjustment. For a given input image, we extract the feature vectors $y_i \in \mathbb{R}^{d_y}$ of the $m$ objects with the highest confidence from the object detection model, transforming these into matrix $Y = [y_1, y_2, ..., y_m] \in$

$\mathbb{R}^{m \times d_y}$. If the number of detected objects in an image is less than m, we applied zero-padding to compensate. To process the input text representation, we augmented it with special positional markers at the beginning and end, designated as [cls] and [sep], respectively. Additionally, we placed [start] and [end] markers before and after the two target entities to indicate their positions. Subsequently, we standardized the length of all descriptions by extending them to the maximum length using the [pad] token. Alongside this, we introduced a token mask composed of zeros and ones, where a '1' represents an actual token, and a '0' signifies the presence of a [pad] token. This mask serves as a record for differentiating between meaningful tokens and padding. To obtain the textual semantic representation, the two sequence are fed into a BERT encoder, and transformed into a matrix $X \in \mathbb{R}^{l \times d_x}$.

Then, we performed an attention mechanism to obtain the semantic alignment weight $\beta$ by equations (5) to (8):

$$K = W_k X + b_k \quad (5)$$

$$Q = W_q Y + b_q \quad (6)$$

$$V = W_v X + b_v \quad (7)$$

$$\beta = softmax(\frac{QK^T}{\sqrt{d}}) \quad (8)$$

Here $W_k, W_q, W_v, b_k, b_q, b_v$ are learnable features, and $d$ is a constant. The overall visual representation, denoted as $\hat{y}$, is derived by summing the elements of $\beta V$ row-wise. Subsequently, the vector corresponding to the two target entities $\hat{v}$, is extracted as [ $v_{[E1_{start}]}$, $v_{[E2_{start}]}$] from matrix $V$, utilizing the [start] and [end] markers. Finally, the output relation distribution is computed as $softmax(MLP(concat(\hat{v}, \hat{y})))$.

*4.3 The generative approaches*

Large Language Models (LLMs) have exhibited exceptional zero-shot reasoning capabilities, which is showcased by their ability to generate detailed rationales as part of the problem-solving process. This proficiency has led to the extensive use of these substantial models in performing natural language inference tasks (Xie et al, 2023; Ma et al, 2023). On the other hand, due to their accumulation of extensive design knowledge and their advanced capacity for pattern recognition and generation, LLMs are becoming pivotal in design-related applications (Wang et al, 2023; Ding et al, 2023). Their ability to interpret and apply intricate design principles enables them to assist in the creative process, offering innovative solutions and enhancing the efficiency and quality of design outcomes. Building on these considerations, we integrated the algorithm described in 3.2 with LLMs, aiming to facilitate the interpretation of combinational pairs.

As depicted in Figure 5, we elicit the chain-of-thought (CoT) from LLM and decompose the task into 3 simple steps. The initial step focuses on the LLM's determination of the base element, from an input image and its corresponding description. This stage concentrates the LLM's attention on identifying a keyword or phrase pivotal to the product's conceptualization, thus eliciting a response that is both precise and confined to the base element. The subsequent phase involves the extraction of all nouns from the product description. In the final phase, the LLM is tasked with analysing the relations between the identified base and the prospective additives. Here, we also provided the category of combinational creativity as an additional hint, adding it to gaining an understanding of the potential relations between the base and the additive.

We mainly implemented this method on OpenAI's GPT-4 and LLaVA (Liu et al, 2023). Both demonstrated comparable performance across various tasks; however, the latter is notably smaller in size and is open-source. For GPT-4, we utilized 'GPT-4 Turbo with vision' for image understanding and 'GPT-4 Turbo' for question reasoning. For LLaVA, we tested both 7B and 13B versions.

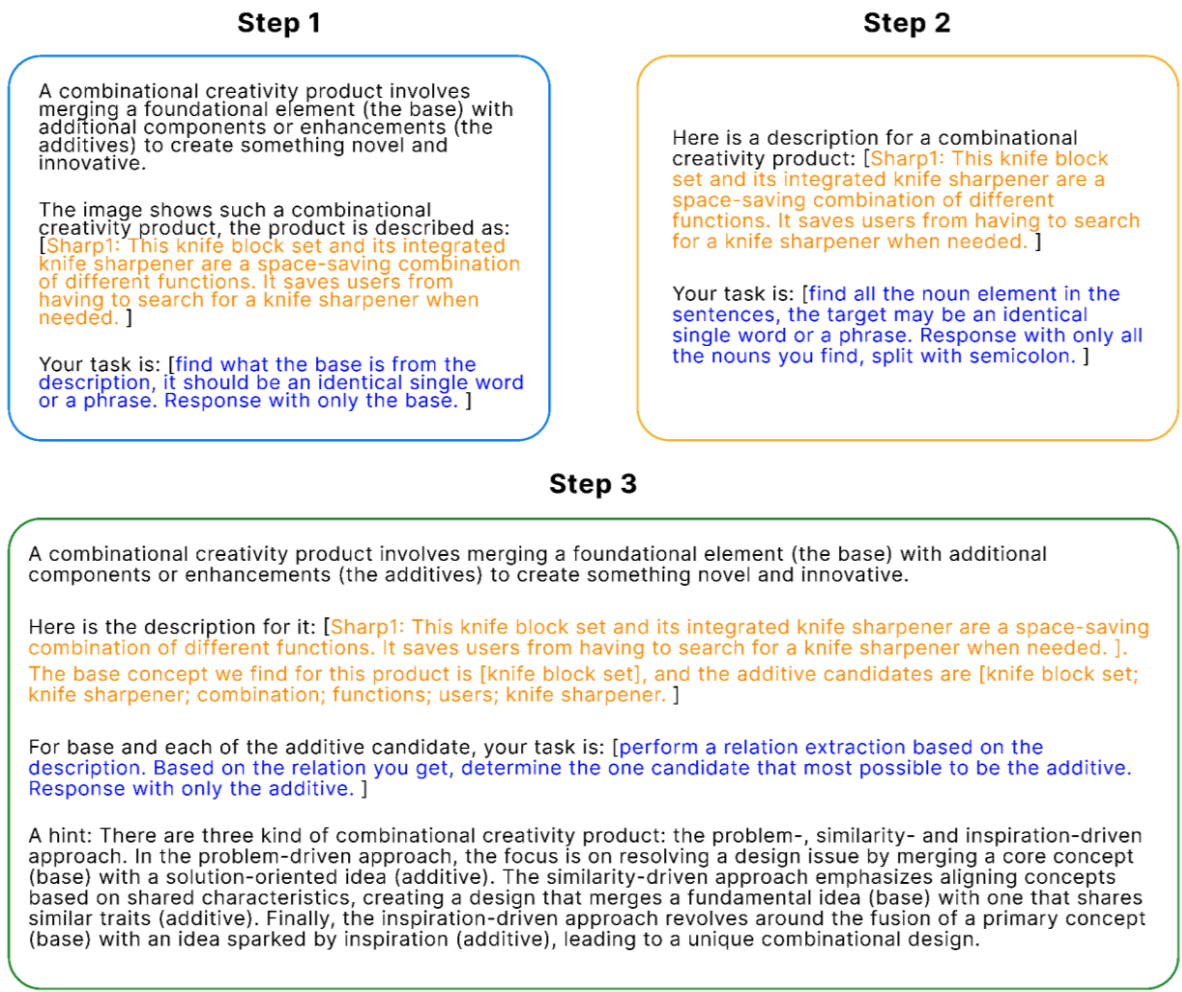

Figure 5. The prompt of proposed LLM-method

## 4.4 The baselines for benchmark

To investigate the effectiveness of our algorithm and to provide a basis for comparison with discriminative methods and generative methods, we implemented a baseline approach based on LLMs and relied solely on its logical reasoning capabilities. The prompting details are shown in Figure 6, we introduced only the basic concept of combinational creativity and the task requirements for extracting combination pairs. To ensure the model's comprehension of the objective, we included two in-context examples. For implementation, we utilized both

'GPT-4 Turbo' and 'GPT-3.5 Turbo' for GPT-series and the 7B and 13B versions for LLaVA.

> A combinational creativity product involves merging a foundational element (the base) with additional components or enhancements (the additives) to create something novel and innovative.
>
> You task is: [decide the base and additive element from the material I give you.]
>
> Example 1 :
> Input [Baby Bottle Drying Rack : The form is inspired by a natural tree shape and eliminates water pooling and prevents minerals and bacteria from building up.]
> Output [Base: Drying Rack; Additive: Tree]
>
> Example 2 :
> Input [Sharp 1 : This knife block set and its integrated knife sharpener are a space-saving combination of different functions. It saves users from having to search for a knife sharpener when needed.]
> Output [Base: Knife Block ; Additive: Knife Sharpener]
>
> The current material is: [ Yedoo Wolfer: The Yedoo Wolfer racing scooter features an attractive appearance and very good riding characteristics, which are similar to a bicycle. ]
>
> Answer like the example output, both base and additive can be found in the material.

Figure 6. The prompt of vanilla LLM-method

## 5 Experiment and Results

### 5.1 Experiment

To assess the accuracy of these interpretations, we applied two criteria to verify the correctness of the predicted bases and additives. Firstly, it should be identical if the base or additive contains only one word. Secondly, at least one keyword must be identical if the base or additive contains a phrase. Using product sample 1 in Table 2 as an example, the predicted additive must include the word 'tree'. The predicted base must contain the word 'rack' at least, as 'rack' is involved in the keyword 'drying rack'.

In the unimodal method approach, clarifai generates the top-10 predictions. Each of these predictions is then paired with noun entities identified by spaCy. We calculate the similarity for each pair, and the noun entity yielding the highest similarity value is designated as the predicted 'base'. In contrast, the multimodal approach utilizes CLIP for creating mappings between text and images. This method directly yields a noun that is predicted as the

'base', thereby obviating the necessity for further similarity computations and retrieval processes. In the process of relation extraction, both methods provide a specific predicted relation, for example, 'part of'. We conducted a similarity analysis by comparing these predicted relationships against our predefined set of relations. The noun that aligns most closely, as indicated by the highest similarity score, is chosen as the predicted 'additive'. Unlike the discriminative approach, in the generative methods, the inference processes are internally handled by the LLMs. Their output consists of two noun entities from the textual description, which are then designated as the 'base' and 'additive' respectively.

*5.2 The overall result for combinational creativity interpretation*

As illustrated in Table 3, we reported on four metrics. The metric 'Both' indicates cases where both the base and additive were correctly predicted. Conversely, 'None' refers to instances where both predictions were incorrect. For 'Base' and 'Additive', they offer the performance of individual predictions, respectively. To encapsulate the overall impression: our multimodal method demonstrated superior overall accuracy, achieving a 72% success rate, which outperformed the GPT-4 method by a margin of 2%. In terms of minimizing completely incorrect predictions, however, GPT-4 led the way with only 3% of samples categorized as entirely wrong, narrowly besting the multimodal method, which registered 4.5% in this metric.

Table 3. Results of combinational creativity interpretation

| Type | Method | Overall measure | | Single measure | |
|---|---|---|---|---|---|
| | | Both↑ | None↓ | Base↑ | Additive↑ |
| Discriminative | The unimodal | 48% | 27% | 69% | 52% |
| | The multimodal | **72%** | 4.5% | 87.5% | **80%** |
| Generative | GPT-4 | 70% | **3%** | **94.5%** | 72.5% |
| | LLaVA-7B | 45.5% | 26% | 60% | 59.5% |
| | LLaVA-13B | 62% | 15% | 83% | 64% |
| Baseline | Vanilla GPT-3.5 | 54.5% | 23% | 73.5% | 58% |
| | Vanilla GPT-4 | 64% | 10.5% | 84.5% | 69% |
| | Vanilla LLaVA-7B | 34% | 43.5% | 43.5% | 47% |
| | Vanilla LLaVA-13B | 41% | 35% | 55% | 51% |

Our first insight was derived from comparing generative methods against baselines, such as GPT-4 versus vanilla GPT-4. Here, 'vanilla' refers to the utilization of LLMs inherent reasoning capabilities without any external algorithm aid, as described in Section 4.4. Our method improved accuracy in interpreting combinational creativity by 6% for GPT-4, 11.5% for LLaVA-7B, and 21% for LLaVA-13B, while reducing completely incorrect rates by 7.5%, 17.5%, and 20% for each model respectively. We also endeavored to comprehend the source of these improvements. To achieve this, we interacted with the LLMs, inquiring how they understand the task of interpreting combinational creativity and the methods they employ to tackle it. As a result, we discovered that LLMs primarily rely on functionality as a clue to understand combinational creativity. They determine the base by analysing the main function of the product, while the additive is identified as the concept that provides the most additional functionality to the product. This notion aligns fundamentally with the problem-driven approach to combinational creativity; however, when it comes to similarity-driven (Figure 7 (a)) and inspiration-driven (Figure 7 (b)) approaches, it falls short because combinations based on similarity or inspiration do not always require a narrative driven by functionality. Our approach resolves this limitation by integrating supplementary contextual cues and heuristic analysis, as well as employing relational back-inference. When compared to the performance of the baseline and generative methods on identifying additives, our method achieved an enhancement of 3.5% for GPT-4, 12.5% for LLaVA-7B, and 13% for LLaVA-13B.

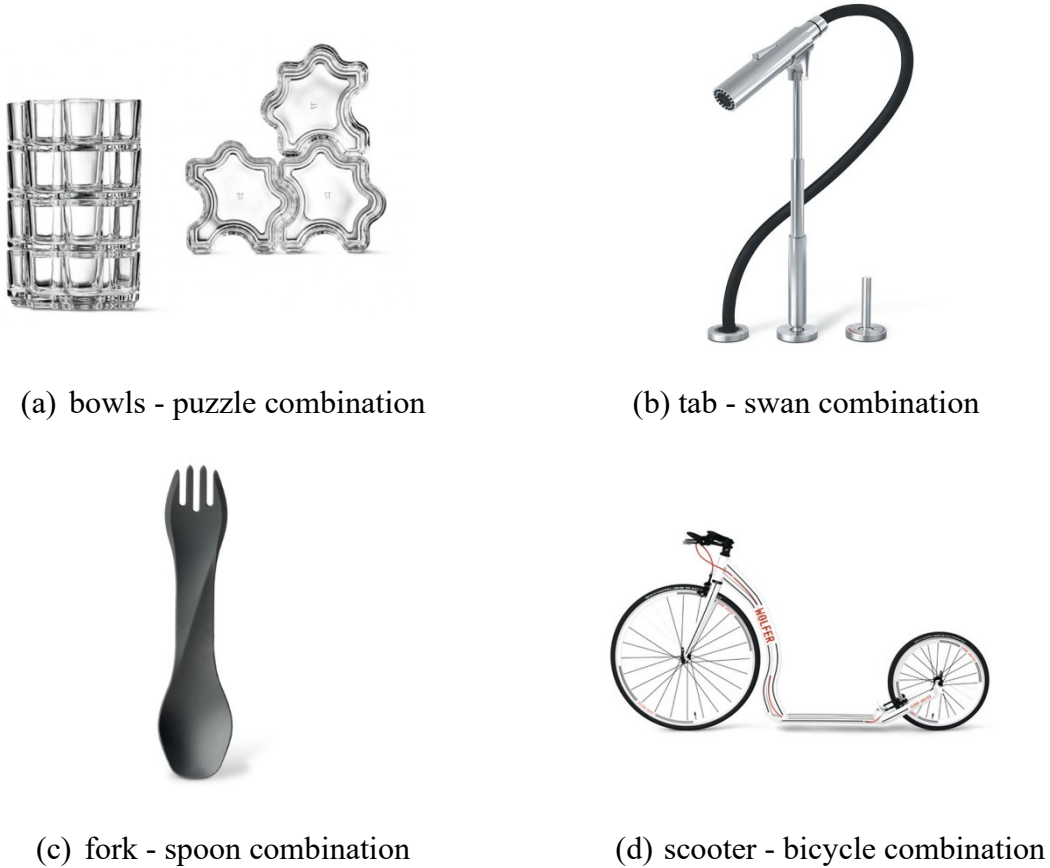

(a) bowls - puzzle combination  (b) tab - swan combination

(c) fork - spoon combination  (d) scooter - bicycle combination

Figure 7. Examples of error case

Our second insight emerged from analysing single measures, revealing that in the context of combinational creativity interpretations, detecting the additive is consistently more challenging than identifying the base, across both our method and the baseline approaches. Statistically, the largest discrepancy was observed in GPT-4, where the accuracy for the base exceeded that of the additive by a significant margin of 22%. The primary reason for this lies in the fact that the base forms the main part of the product, making it generally more identifiable in both images and text. In contrast, the additive tends to be less conspicuous and more subtly integrated. By employing relation extraction techniques, we uncover more hidden additives. The last insight arises from an analysis of the error cases. Taking results from the multimodal method as an example, out of 56 erroneous cases, 29 exhibited a reversal in the identification of the base and additive. A primary reason for this is the high prominence of the additive in some products. Typically, additives are minor components, like the knife block in Table 2, where the Sharpener only takes a small portion. While sometimes they may even constitute up to 50% of the product, such as the fork – spoon combination and the scooter – bicycle combination shown in Figure 7 (c) & (d). In this scenario, a mere qualitative estimation is utilized for interpreting the product image, which may lead to classification errors by clarifai, misalignments in text-image pairing by CLIP, and misunderstandings in image comprehension by LLMs.

*5.2 Modular analysis*

Table 4. Results of modular analysis

| Method | Modular | Correct number | Accuracy |
| --- | --- | --- | --- |
| The unimodal | clarifai | 138 / 200 | 69 % |
|  | spaCy | 200 / 200 | 100 % |
|  | context-aware RE | 152 / 200 | 76 % |
| The multimodal | CLIP | 175 / 200 | 87.5 % |
|  | spaCy | 200 / 200 | 100 % |
|  | MEGA | 163 / 200 | 81.5 % |

In this section, we present a modular analysis of the discriminative method, as shown in Table 4. For the interpretation of the base in unimodal methods, we utilized the general image recognition model of clarifai. We analysed the top 10 categories with the highest probabilities from the model's output. If one of these categories matches the base, we consider it as identified. We employed spaCy's similarity module to assess the similarity between the two concepts. For example, the similarity between 'bulb' and 'lamp' is 0.781, while between

'bulb' and 'fire' it is 0.243. If the similarity score between the model's output and the base exceeds 0.75, we consider them to be a match. As a result, the clarifai successfully detected 138 bases. In contrast, the multimodal method leveraged CLIP for base interpretation by integrating textual analysis. Here, we used the top-1 result as the base, successfully identifying 175 bases.

In the second step, both methods employed spaCy's NER module to extract all noun entities from the textual description. Since the aim of this step is to identify all potential additive candidates, if the additive appears among the extracted entities, we consider it correctly identified. As a result, spaCy performed the noun extraction task flawlessly with 100% accuracy. Although this approach could bring noises, such as the entities extracted other than the additives, it is guaranteed that all additive candidates are captured without missing extractions for the following processes. Further research is required to explore potential methods for filtering the noises to increase the accuracy of the downstream processes. A potential method could be topic extraction which extracts topics from contextual data by employing TF*PDF or TF-IDF (Qaiser and Ramsha, 2018; Gomes et al, 2023).

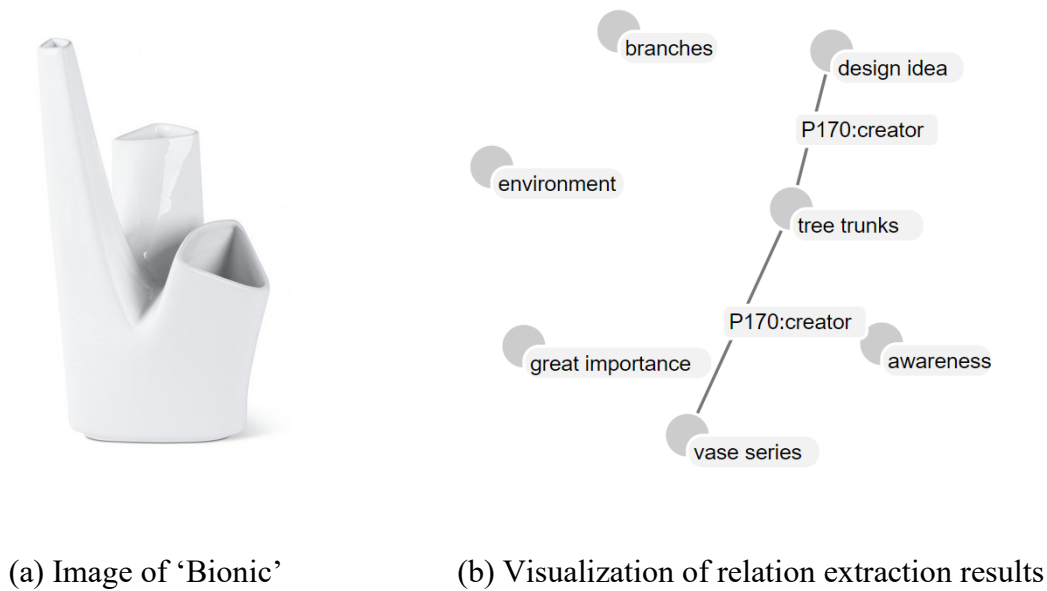

(a) Image of 'Bionic'  (b) Visualization of relation extraction results

Figure 8. An example of relation extraction

The RE modules has reached a 76% accuracy for context-aware RE in the unimodal method and 81.5% accuracy for MEGA for the multimodal method with regards to the extractions of combination pairs. In this test, due to the absence of base identification in the first step, we opted to randomly select candidates from the recognized entities to serve as both base and additive candidates. For instance, if there are $n$ entities, we would test $C_n^2$ different combinations. We determined the final base-additive combinations through predefined relations shown in Table 1. Since some relations lack directionality, we treated reversed results of the base and additive also as correct predictions. As an example, we present the image and relation recognition results of 'Bionic' in Figure 8, with its product description as follows:

*Bionic: the design idea for this vase series was inspired by tree trunks and their branches, and aims to increase awareness of the great importance of preserving the environment.*

From the relation extraction, it can be observed that a total of 7 entities have been identified in the description of Bionic. A notable observation is that most of these entities do not have relationships with each other. This is a common scenario, which can help simplify the decision-making process. In this test, two relations are extracted: vase series – tree trunks and tree trunks – design idea. In the complete method, once vase series is determined as the base by image interpreting, we can easily arrive at the correct answer, even without the need to assess the similarity of relations. However, there also exist several challenging scenarios and failure cases:

- The target base or additive of a combination design does not exist in the textual description but only in the name of the product. For example, part of the name of the product sample 1 in Table 1 is the base, while the base does not exist in the textual descriptions. Even though the product name and corresponding textual descriptions are delivered into the RE module together, it is challenging to detect the relation between the base and additive because of the low connection and appearance frequency.

- The target base and additive of a combination design exist in different sentences. Although there is a limitation regarding the number of sentences for describing the design, it is possible that the base and additive could appear in different sentences. This could result in missing extractions, and a scenario where the base and additive are indirectly connected via an intermediary entity. Further studies are required to solve the second situation, while the first one is fatal for the final identification.

- The target base or additive of a combinational design is extracted in a relation while the other is not. For some designs, their names and bases are mixed for describing them. This might confuse the RE module in relation extraction. In some other designs, the additives appear together with other entities, which might disturb the relation extraction. Although this case is tackled in the verification stage, the ranking of entities may not guarantee the accuracy of the extraction.

### *5.3 Ablation study on the role of image*

In this section, we explored the role of product images in interpreting combinational creativity. The motivation behind this inquiry stems from the fact that, in the real world, textual descriptions of products are relatively abundant and easily accessible, whereas images that accurately convey design concepts are more valuable. If our method can still achieve satisfactory interpretive results in the absence of images, it would have greater applicability and utility. We conducted tests on generative methods and the baselines. For the generative approach, in the absence of images, we prompted the LLMs to use the identified noun entities as candidates for both base and additive as we did when testing the RE module in Section 5.2.

As for the baselines, we maintained the prompt presented in Section 4.4 but did not provide any image input.

Table 5. Results of ablation study on the role of image

| Method | Overall measure | | Single measure | |
| --- | --- | --- | --- | --- |
| | Both↑ | None↓ | Base↑ | Additive↑ |
| GPT-4 w/o image | 65% | 5% | 88.5% | 71.5% |
| LLaVA-7B w/o image | 40.5% | 35.5% | 54.5% | 50.5% |
| LLaVA-13B w/o image | 58% | 23% | 69% | 66% |
| Vanilla GPT-4 w/o image | 62% | 14% | 76.5% | 71.5% |
| Vanilla LLaVA-7B w/o image | 31.5% | 47% | 43% | 41.5% |
| Vanilla LLaVA-13B w/o image | 40% | 37% | 51.5% | 51.5% |

We presented all results in Table 5. Overall, images contribute positively to the interpretive performance of our methods. For our generative approach, we observed a decrease of 5 % for GPT-4 and LLaVA-7B, and 4 % for LLaVA-13B in both correct situations when images were not included. The baseline group also experienced a similar trend, but with a smaller loss in accuracy: 2%, 2.5%, and 1% respectively. We hypothesize that our strategy of having the LLMs actively examine images to determine the base might have contributed to this effect. When we focus on single measures, the disparity in predicting the base and additive still exists, but the gap has been narrowed. Interestingly, in the absence of images, LLaVA-13B (66%) and vanilla GPT-4 (71.5%) actually performed better in identifying additives than before (64% and 69%, respectively). A possible reason for this improvement could be that by treating base and additive equally in relation extraction, we eliminated instances where base and additive were previously predicted in reverse. In the earlier process, where the base was determined through images, the close proportion of base and additive (such as the fork – spoon combination in Figure 7 (c)) often led to a complete loss of interpretation for that sample.

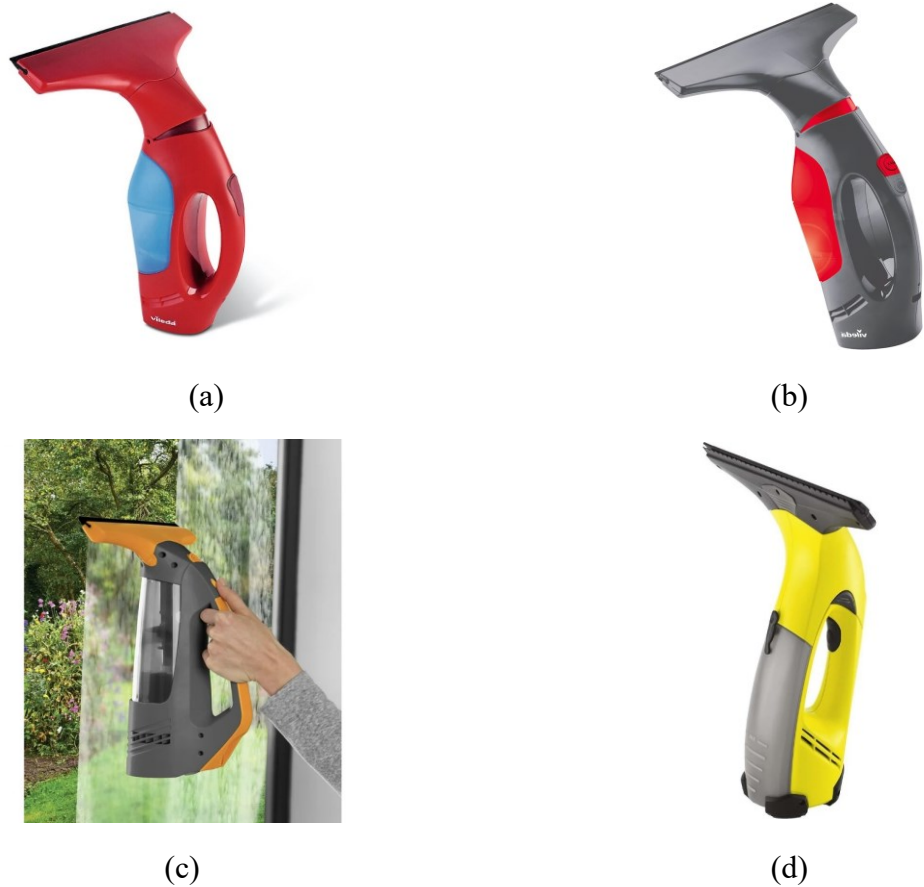

Figure 9. Different images for one combinational creativity product

On the other hand, we are curious whether increasing the number of image representations can resolve the previously reported missing or confused descriptions of the 'base' and 'additive' in the text. Since the original dataset contains only one image for each combinational creativity case, we utilized the similar image search features of browsers and shopping websites to acquire additional images. In total, we selected 10 cases for testing from the errors made by our multimodal approach. These included scenarios where the prediction of either the 'base' or the 'additive' was incorrect, both were incorrect, or the predictions were reversed. We show one of our collected cases as Figure 9. We ensured that the acquired product images are consistent with the textual descriptions of their functions. Additionally, we strived to make the new images exhibit characteristics different from the original image (a). In this case, (b) features a different color, (c) further describes its usage scenario, and (d) differs in appearance and angle. For our multimodal approach, the prediction of 'base' indeed improved, as three out of the five originally incorrect 'bases' were successfully identified. However, the reversal of 'base' and 'additive' remains the most severe error and the same issue also occurred with GPT-4. In the tests with GPT-4, we attempted to understand how additional images affect its decision-making. Overall, GPT-4 was more stubborn than our multimodal approach, often maintaining 'My answer remains consistent.' even when its predictions were incorrect.

## 6 Discussion

### *6.1 Bottleneck of the integrated interpretation algorithm*

The main bottleneck of the approach, in terms of the module performance, is the image interpretation module. Clarifai and CLIP, employed in this study, are trained on datasets primarily consisting of conventional images, capturing typical and standard patterns of specific object categories. However, creativity usually involves deviation from these norms. This deviation can challenge the models, potentially leading to difficulties in accurately interpreting the unique and unconventional aspects that are characteristic of creative products. A potential solution to this could be that training them by using an intensive product image dataset on top of the pre-trained model by transfer learning (Shin et al., 2016). Besides, according to the analysis of the factors influencing the accuracy rates discussed in the preceding section, solutions that can alleviate their negative impact could be provided. For instance, increasing the number of images of a combinational design from various views (Liu et al, 2018); delivering the invisible feature issue to the RE module with extra attention (Feng et al, 2023); employing finer-grained methods to interpret images which are capable of distinguishing specific differences between similar categories (Wei et al, 2021).

In terms of the extraction of combination pairs, additive extraction is considered to be another bottleneck. As discussed in the preceding sections, features from the additive tend to be employed less in a combinational design in comparison with features from the base. This is the main reason that leads to a low combination pair extraction accuracy. Several potential improvements could be considered:

- The name of a combinational design needs more attention. The entities involved in the name of the design have a high possibility to contain the base or/and the additive, if the name is not too fancy to reflect the essence.
- Choose a Relation Extraction (RE) model architecture capable of multi-sentence inference. An intermediary which connects the base and additive, while they appear in different sentences, of a combinational design could thereby be analysed.

*6.2 Implications of the study*

In this study, we discussed the basic interpretation of combination elements (base and additive) from conceptual designs, and analysed the difficulty of this task from a computational perspective. Compared to most works in the field of conceptual design, such as bisociative knowledge discovery (Ahmed et al, 2018; Zuo et al, 2022) and visual conceptual blending (Han et al, 2018; Wang et al, 2023), which focus on the transformation from design data to design creativity, our work proposes a preliminary approach to interpret the process from design creativity back to data. Such a bi-directional transform between data and creativity can form a closed creative knowledge reuse loop so that creativity is not only produced from existing mechanisms in creative systems but also benefits from produced creative knowledge with creativity interpretation mechanisms. From a long-term perspective, this bi-directional transform enables a creative system to evolve from such a data-driven cycle, thus achieving continuous creativity. In this sense, our work represents an important complement to data-driven design and leads to continuous creativity in conceptual design. Furthermore, from a practical design standpoint, understanding the underlying structure and relationships in combinational creativity empowers designers to refine their methods of integrating diverse elements into a product. By grasping the structure and relationships in combinational creativity, designers can better evaluate the effectiveness of their designs. For novice designers, this approach has educational benefits as well. It can serve as a foundational tool for teaching design principles, allowing them to recognize and apply creative combinations effectively.

# 7 Conclusion

This study explores how to interpret a combinational creative design by extracting the base and additive as a combination pair from corresponding image and textual descriptions in an AI-based approach. A heuristic interpretation algorithm is proposed in this study to extract the combination pairs jointly. Based on this algorithm, we have explored approaches utilizing both discriminative and generative AI models. By conducting experiments on a combinational design creativity dataset, it is shown that our proposed interpretation approaches could successfully extract combination pairs, especially the bases, from real-world combinational creative designs. However, it is also found that additives are more challenging to be extracted comparing with the bases. Factors which might have caused the issue are discussed in the study. Observations and potential improvements for the interpretation approach are also discussed in the study.

# Acknowledgements

This research was funded by National Key R&D Program of China (Grant No. 2022YFB3303304), and National Natural Science Foundation of China (Grant No. 62207023).